\documentclass[letterpaper, 10 pt, conference]{ieeeconf}  % Comment this line out if you need a4paper

\IEEEoverridecommandlockouts                     % This command is only needed if 
\usepackage{graphics} % for pdf, bitmapped graphics files
\usepackage{epsfig} % for postscript graphics files
\usepackage{mathptmx} % assumes new font selection scheme installed
\usepackage{times} % assumes new font selection scheme installed
\usepackage{amsmath} % assumes amsmath package installed
\usepackage{amssymb}  % assumes amsmath package installed
\usepackage{hyperref}
\usepackage{subcaption}
\usepackage{xcolor}
\usepackage{multirow}
\usepackage{array}
\usepackage{adjustbox}

\usepackage[bottom]{footmisc}
\title{\LARGE \bf
SemanticSLAM: Learning based Semantic Map Construction and Robust Camera Localization
}

\author{Mingyang Li, Yue Ma, and Qinru Qiu\\Department of Engineering and Computer Science, Syracuse University\\ \{mli170, yma183, qiqiu\}@syr.edu}

\begin{document}

\maketitle
\thispagestyle{empty}
\pagestyle{empty}

%%%%%%%%%%%%%%%%%%%%%%%%%%%%%%%%%%%%%%%%%%%%%%%%%%%%%%%%%%%%%%%%%%%%%%%%%%%%%%%%
\begin{abstract}

Current techniques in Visual Simultaneous Localization and Mapping (VSLAM) estimate  camera displacement by comparing image features of consecutive scenes. These algorithms depend on scene continuity, hence requires frequent camera inputs. However, processing images frequently can lead to significant memory usage and computation overhead. In this study, we introduce SemanticSLAM, an end-to-end visual-inertial odometry system that utilizes semantic features extracted from an RGB-D sensor. This approach enables the creation of a semantic map of the environment and ensures reliable camera localization. SemanticSLAM is scene-agnostic, which means it doesn't require retraining for different environments. It operates effectively in indoor settings, even with infrequent camera input, without prior knowledge. The strength of SemanticSLAM lies in its ability to gradually refine the semantic map and improve pose estimation. This is achieved by a convolutional long-short-term-memory (ConvLSTM) network, trained to correct errors during map construction. Compared to existing VSLAM algorithms, SemanticSLAM improves pose estimation by 17\%. The resulting semantic map provides interpretable information about the environment and can be easily applied to various downstream tasks, such as path planning, obstacle avoidance, and robot navigation. The code will be publicly available at \url{https://github.com/Leomingyangli/SemanticSLAM} 

\end{abstract}

\begin{figure*}[!t]
\centering
\includegraphics [width=0.9\textwidth]{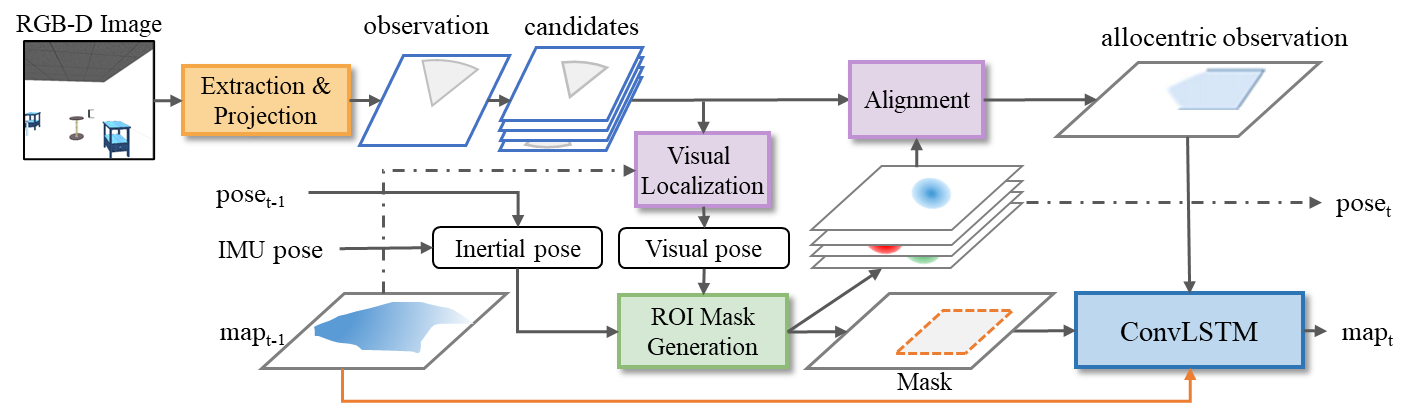}
\caption{\centering \textbf{System Overview of SemanticSLAM}}
\label{fig:network}
\vspace{-0.5cm}
\end{figure*}
%%%%%%%%%%%%%%%%%%%%%%%%%%%%%%%%%%%%%%%%%%%%%%%%%%%%%%%%%%%%%%%%%%%%%%%%%%%%%%%%
\section{INTRODUCTION}

Visual Simultaneous Localization and Mapping (VSLAM) is a challenging computational problem that has wide application areas in various fields such as robotics and autonomous vehicles. The goal of VSLAM is to construct a map of an unknown environment using visual input while simultaneously estimate the position and orientation of the camera.
% \cite{SLAM}

Traditional VSLAM techniques, such as ORBSLAM \cite{orbslam2}, perform pose estimation by comparing matched keypoints of camera inputs with recent keyframes. The algorithm first extracts keypoints, which are image features selected either based on predesigned filters \cite{orbslam2,dynaslam} or machine learned models\cite{liftslam},  from the camera input and track them in keyframes by matching the descriptors. The pose estimation is then performed by minimizing the reprojection error between the matched 3D map points in world coordinates and 2D keypoints. A considerable large number of matched keypoints on consecutive frames is required to reliably perform pose estimation, therefore, the scene continuity in adjacent frames is important. The camera images must be captured and processed frequently to ensure a high level of similarity among the consecutive frames. However, frequent image processing and keypoint comparison will cause significant computation overhead and to record all keyframes of the environment will result in a high storage complexity. 

Rather than memorizing the image features of the visual input, human navigates in an environment by roughly estimating their position based on relative location of surrounding objects. Neuroscience research shows that the place cell, i.e., the set of neurons associated with locations, located in the hippocampus and involves in the function of episodic and semantic memory \cite{eichenbaum1999hippocampus, herweg2018spatial}. In this work, we present SemanticSLAM. The algorithm performs localization and map construction using extracted visual semantic information. The map is a 2D array of neuron clusters, each representing a grid area in the environment. The state of the neuron clusters, which can be written as a vector, represents the semantic information of the corresponding grid area. By comparing the observation and the semantic map, a rough estimation can be made about the camera pose. With the estimated pose and the observed semantic information of the surrounding area, the algorithm will update the map to include the new observations. 

The pose estimation described above may have large error at the early phase of a mission when the map has not been constructed. Because the robot has no prior knowledge of the environment, no matching could be found by comparing the observation with the map. In addition, the observation is not always perfect. For example, the most common observation error is due to obstruction. Using imperfect observation to update the map will also introduce errors. In this work, a convolutional long-short-term-memory (ConvLSTM) is trained to correct errors during map update such that the errors will eventually converge instead of being magnified with the iteration. To improve the accuracy of pose estimation at the beginning of the mission,  we will leverage the reading from a low cost Inertial Measurement Unit (IMU) to cross check the pose and narrow down the region of the map update. Solely rely on the IMU is not an option for pose estimation as the intrinsic error will accumulate and eventually become unbearable. However, using the IMU input to bootstrap the pose estimation at the beginning of the mission is viable.

Compared to traditional SLAM, the SemanticSLAM's advantage is two folds. Firstly, it does not require high frequency observation and image processing. While different distance, view angle and luminance level may change image features in the observation, after the semantic extraction, such low level variance will be filtered out. Therefore, continuity in the view point is not a necessary condition to find a semantic match. Secondly, compared with maps storing image features and keyframes, a map with semantic information requires much less memory and is more human interpretable. The semantic map can be directly used for mission and navigation planning and can be easily shared among robots as well.

We evaluate the effectiveness of SemanticSLAM, on a dataset obtained from the Gazebo \cite{gazebo} simulation platform using RGB-D cameras and IMU data. Our experimental results demonstrate that SemanticSLAM outperforms other representative methods in terms of accuracy and adaptability to new environments. 
% Furthermore, we also show how each suggested methodology contributes to enhance the overall efficacy of the system. 
% The remainder structure of the paper is as follows: Section II provides a literature review of related works in this field, Section III presents our proposed architecture and algorithm, Section IV describes our dataset and experimental setup, and presents the results and analysis. Finally, in Section V, we conclude the paper and discuss potential directions for future research.

\section{Related Work}
Existing VSLAM techniques, such as ORB-SLAM \cite{orbslam2} and LIFT-SLAM \cite{liftslam}, employ parallel threads for tracking, mapping, relocalization, and loop closing. Traditionally, feature points are extracted from images using predefined descriptors such as SIFT \cite{sift}. The map is a collection of the frames consists of extracted feature points. Camera pose is estimated by minimizing the reprojection error between the query frame and the stored frames that have the largest number of matched feature points. To ensure that an abundance of map points can be found, large amount of keyframes must be stored in the map, and the observation must have a significant amount of continuity. Hence these methods often require significant computational and memory resources.  After pose estimation, some recent works \cite{maskfusion,neuralslam} also extract semantic features from the camera image and built a semantic map. However, the semantic map is not involved in the process of pose estimation. \par
% cnnslam,

There is also a class of research work that performs visual localization  without map. One of the techniques is relative camera pose regression (RPR).  The algorithm directly predict the displacement of the camera position between a reference image and a query image using neural networks \cite{superglue, deepvo}. Such method typically requires highly correlated images, hence needs to be performed frequently. Under low frame rate, the RPR will suffer from accumulated error and eventually lose tracking. Because the RPR does not perform loop closure, additional processing is needed to prevent errors from accumulating.  A lightly different technique is absolute pose regression(APR), which approximates the relationship between a query image  and reference scene using a neural network. \cite{posenet,lessismore}. APR is scene-specific.  A new neural network model must be trained for a new application scenario. \par

To improve the accuracy and robustness of visual localization, the Visual-Inertial Localization method has been introduced. Classical feature-based methods, such as ORBSLAM3 \cite{orbslam3}, utilize IMU data to provide a metric pose estimate, which acts as a constraint for visual localization. On the other hand, there are deep learning-based approaches that combine\cite{vinet} or fuse\cite{selfvio} learned features from both sequential images and IMU data to perform visual inertial odometry. These methods leverage the power of deep learning techniques to effectively integrate visual and inertial information, resulting in improved visual inertial odometry performance. However, these methods are susceptible to various IMU sensor noises that accumulate drift errors, leading to inaccurate pose estimation as the agent moves. \par

Instead of relying on the matching of keyframes or comparing between the query and reference frames, we project the observed image to build an observation map and try to correlate it with an allocentric "global" map that has been constructed from past observations for pose estimation. To the best of our knowledge, the only work that adopts a similar approach is the MapNet\cite{mapnet}. However, MapNet stores image features instead of semantic features of the environment in the global map. Hence the map cannot be directly used to guide the mission planning. Furthermore, varying image features for the same object may be received when observed from different angles, which makes it difficult to find a match between the current observation and past observations stored in the map.

\section{Method}
In this section, we present the SemanticSLAM. The structure of the framework is illustrated in Figure \ref{fig:network}. It receives the RGB-D observation of the environment and provides an estimated pose of the camera and a semantic map.

We consider the camera localization during navigation. The environment is considered a $H \times W$ grid map, each grid is a possible camera location. The potential camera orientations are discretized into $R$ levels. The robot observes the environment and estimates its location at discrete time $t$, the time intervals between adjacent observations and estimations does not have to be equal. For each step $t$, the output of SemanticSLAM is an array of probability values, denoted as $\mathbf{p_t}\in [0,1]^{R\times H \times W}$, and $\sum_{r,x,y}^{R,H,W}p_t^{r,x,y}=1$, $\forall t$. The 
$(r, x, y)$th entry of the array gives the probability that the camera is at location $(x,y)$ and its orientation falls into the $rth$ level. The index of the entry with the largest value is the estimated pose $\bar{p}_t=argmax_{x,y,r}{p_t^{r,x,y}}$. The framework also constructs and maintains a neural-symbolic map $m_t \in {[0,1]}^{L \times H \times W }$ of the environment, where $L$ is the dimension of the semantic feature vector. 

The input of the system includes the information $(I_t,d_t)$ from RGB-D camera reading, where $I_t \in \mathbb{R}^{s_1 \times s_2 \times 3}$ is the set of pixels in  RGB image and $d_t \in \mathbb{R}^{s_1 \times s_2\times 3}$ gives the 3D point cloud of each pixel in the RGB image. The IMU sensor reading is represented as a triplet,  $u_t = [\Delta x_t,\Delta y_t,\Delta \theta_t]$, which describes IMU measured position and rotation changes from last observation. 

% \begin{figure}[t]
% \centerline{\includegraphics [width=0.35\textwidth]{./images/yolo.png}}
% \caption{\textbf{An illustration of Semantic Observation Extraction and Projection}}
% \label{fig:yolo}
% % \vspace{0in}
% \end{figure}

\subsection{Semantic Feature Extraction and Projection}\label{sec:v_noise}
In this work, we use the class labels of the foreground objects as the semantic feature of the environment. These semantic features are detected from the RGB image, then projected onto a 2D observation map based on the depth information. 
% The process is illustrated in Figure \ref{fig:yolo} 

\textbf{Object Extraction}. We detect the foreground objects from the received RGB image $I_t$ using a pre-trained Yolo\cite{yolov3} model, which gives a bounding box for each detected object. Next, the object is separated from its background using a semantic segmentation model SAM \cite{SegAnything}. After object extraction, we received a group of $N$ foreground objects, $\{(A_i,C_i),{{0} \leq {i} \leq {N-1}}\}$, which $A_i$ is the set of pixels in the RGB image, and $C$ is its class label, which will be used as the semantic feature in our map.  

\textbf{Feature Projection}. Using the one-to-one pixel correspondence between the depth image ($d_t$) and the RGB image ($I_t$), the semantic features of the foreground object will be projected onto an egocentric(i.e., using camera coordinate system) 2D observation map $o_t\in \mathbb{R}^{L \times h \times h}$. The observation map represents a top-down view of an  ${h\times h}$ region centered at the camera location with orientation 0. Associated to each map location is an $L$ dimensional vector, storing the semantic features at that location. For each pixel $a \in {A_i}$ of the $i$th foreground object, we first identify its corresponding $(x,y)$ location is using the depth image,then increment the value $o(C_i, x, y)$ by $1/|A_i|$, where $C_i$ is the class label of the object and $|A_i|$ is the cardinally of the pixel set $A_i$. At the end of the feature projection process, an entry $o(l, x, y)$ can be written as the following:

%utilizing the point cloud $d_t$. Typically, we convert each semantic pixel into a $L$-dimension one-hot vector, and project it onto the corresponding position in $o_t\in \mathbb{R}^{L\times h\times h}$ based on pointcloud $d_t$. However, due to varying image resolutions and observation unit size, the projection is not always a one-to-one mapping where a unique cell in $o_t$ may contain multiple pixels in $I^{\prime}$. To address this issue, we alternatively utilize the ratio of pixel projected on to the entry $(x,y)$ of $o_t$ to total semantic pixels in the bounding box as a measure of confidence in $o_t$, defined as (omitting $t$ for clarity):
\begin{equation}
    \label{eq:projection}
\def\thefootnote{a}\footnotesize
    \begin{aligned}
    o(l,x,y) = \sum_{i,C_i=l} \frac {|a, a\in {A_i} \& d_a(x,y)=(x,y)|}{|A_i|},
        \end{aligned}
\end{equation}
where $d_a(x,y)$ gives the $(x,y)$ position of the pixel $a$ in the depth image. 

% of pixels in a candidate bounding box with category $l$. The spatial projection $P(\cdot)$ function is defined by:
% \begin{equation}
%     \label{eq:projection_condition}
% \def\thefootnote{a}\footnotesize
%     \begin{aligned}
%     x = [\frac{h-1}{2} + \frac{d_{ij}^x}{k}] \\
%     y = [\frac{h-1}{2} - \frac{d_{ij}^y}{k}]
%         \end{aligned}
% \end{equation}

% where $d_{ij}^x$ is the depth of pixel point $(i,j)$, $d_{ij}^y$ its horizontal displacement. $[\cdot]$ denotes integer rounding and $k$ is the unit size of the observation cell.
As we can see, the $o(l, x, y)$ describes how different classes of objects are distributed across the 2D area. The feature value lower than a threshold $\beta$ (e.g., $0.02$) is considered as noise and will be filtered. This projection is by no means perfect. There are many potential sources of errors, the object detection and classification may be wrong, the segmentation may not be perfect, and the feature projection may contain errors. Even if all of these steps are correct, due to obstructions, objects may not be observed or only partially observed. Hence the observation map contains a significant amount of errors. We leverage a ConvLSTM network to correct the errors  during map construction.
% \begin{equation}
%     \label{eq:ceilfloor}
% \def\thefootnote{a}\footnotesize
%     \begin{aligned}
%     o_l^{xy} =
%      \begin{cases}
%         1 & \text{if $o_l^{xy} \geq \beta$ }\\
%         0 & \text{otherwise}
%     \end{cases}
%     \end{aligned}
% \end{equation}

% By doing this, the observation impliciy obtain the In this work, we assume the presence of a sufficient number of detectable and classifiable objects in the scene, which is typically reasonable for indoor environments. 

\subsection{Visual Pose Estimation}
The semantic observation map $o_t \in \mathbb{R}^{L\times h\times h}$ is egocentric. It then undergoes a set of rotations at multiple viewing angles $2\pi r/R$ for $r\in\{1,...,R\}$ using Spatial Transformation\cite{spatialTransformer}, which generates a set of observations $\acute{o}_t \in \mathbb{R}^{R\times L\times h\times h}$, where $R$ is the number of orientation levels. Each candidate represents a hypothesis of the observation map relative to allocentric(world coordinate system) view angle $r$. The correlation between each observation candidate and the global map  $m_{t-1}$ constructed from previous time step is calculated by applying a 2D convolution on $m_{t-1}$ with kernel $\acute{o}_t$ and stride 1:
\begin{equation}
    \label{eq:pose}
\def\thefootnote{a}\footnotesize
    \begin{aligned}
    {v}_t = \sigma (m_{t-1}\;\mbox{*}\;\acute{o}_t)
    \end{aligned}
\end{equation}
where $\sigma()$ represents the softmax function. We pad the global map $m_{t-1}$ such that the output $v_t$ has the dimension ${R} \times {H} \times {W}$. $v_t$ is a visual pose probability field. Its $(r, h, w)$th entry indicates the degree of correlation between the map and the observation if the camera is located at position $(h,w)$ with orientation $r$ relative to the world coordinate system, which also gives the probability distribution of the camera's location and pose. The entry with the largest value gives the visual pose estimation:  $\bar{v}_t=argmax_{x,y,r}{v_t^{r,x,y}}$. 

%To achieve this, we ensure that the map is padded in a manner that aligns the visual pose probability field with the last two dimensions of the map. %Moreover, we introduce a grouped convolution technique inspired by AlexNet\cite{AlexNet}, which ensures that the matching process only operates on the same semantic level of the two inputs. 

%After performing the visual localization, we acquire an allocentric pose estimate in the map coordinate system from the egocentric observation. This estimate, denoted as $\bar{v}_t\in\mathbb{R}^{3}$, corresponds to the index of the maximum correlation value in the probability distribution $v_t$.  

\subsection{Cross-check with Inertial Pose Estimation}
At the beginning of the mission, before the map is constructed, the observation $o_t$ and the global map $m_{t-1}$ correlates poorly and the aforementioned visual pose estimation will have low performance. We propose to bootstrap the pose estimation using IMU sensor data to mitigate the uncertainty.  An alternative inertial pose estimation, $\bar{u}_t \in\mathbb{R}^{3}$, is obtained. Given the pose estimation in previous step $\bar{p}_{t-1}=(r_p, x_p, y_p)$, the location and orientation changes detected by the IMU sensor over the last observation interval $u_t=[\Delta x_t,\Delta y_t,\Delta \theta_t]$, the inertial pose estimation $\bar{u}_t=(r_i, x_i, y_i)$ can be calculated as the following:
\begin{equation}
    \label{eq:inertial pose}
\def\thefootnote{a}\footnotesize
    \begin{aligned}
    \bar{u}_t = T(\bar{p}_{t-1},\hspace{.1cm} u_t)
    \end{aligned}
\end{equation}
where $\bar{p}_{t-1}\in\mathbb{R}^{3}$ is the allocentric pose estimated in last time step and $u_t\in\mathbb{R}^{3}$ is the egocentric pose displacement sensed by IMU. The transformation function $T(\cdot)$combines these two to calculate the resulting allocentric pose. 

It should be noted that the reading of IMU consists of  Gaussian noise and fixed biased noise, which results in drift error in the inertial pose estimation. The error in the IMU reading is relatively small, however, after several steps of accumulation, it will become quite significant. On the other hand, when the visual pose estimation is wrong, the error is usually drastically large because we are guessing randomly without evidence of matched features. However, when matched features are found, the visual pose estimation tends to provide very accuracy results. 

%We introduce a novel approach to address the error discrepancy between the visual pose estimate and the inertial pose estimate. At each time step, we choose either the visual pose $\bar{v}_t$ obtained from visual localization or the inertial pose $\bar{i}_t$ as the current pose estimate $\bar{p}_t$. 
Based on the above observation, the final pose estimation $\bar{p}_{t}$ will be chosen from either visual estimation $\bar{v}_{t}$ or innertial estimation $\bar{u}_{t}$ based on the following threshold function:
\begin{equation}
    \label{eq:threshold function}
\def\thefootnote{a}\footnotesize
    \begin{aligned}
    \bar{p}_t = 
     \begin{cases}
        \bar{v}_t & \text{if ${\| {\bar{v}_{t,k}-\bar{u}_{t,k}} \|}^{k=2,3} < {\gamma}_1$ }\\ & \text{and ${| {\bar{v}_{t,k}-\bar{u}_{t,k}} |}^{k=1} < {\gamma}_2$ } \\
        \bar{u}_t & \text{otherwise}
    \end{cases}
    \end{aligned}
\end{equation}
where $\gamma_1, \gamma_2$ are the maximum noise that the IMU may have during one sample interval.  If the difference between the visual and inertial estimations is greater than the maximum possible noise level, we assume that the visual estimation is wrong and accept the inertial estimation. Otherwise, we accept the visual estimation result. 
The visual-inertial cross check imposes a bound to the pose estimation error, particularly at the onset of the training phase. Hence it accelerates the convergence of the training. It is also easy to implement as the visual and inertial pose estimation are two parallel and independent threads.
%Additionally, this threshold function operates as an independent parallel thread, allowing it to seamlessly detach from the main model without necessitating any further modifications. Consequently, our model exhibits flexibility, capable of operating as either visual or visual-inertial, as required.}

\subsection{Map Update}
\textbf{Observation Projection}. To update the global map using the observed information, first we need to project the egocentric observations to the allocentric global map at the estimated pose. In visual pose estimation, the pose has a probability distribution $p_t$, which is a 3D tensor with dimensions $R\times H \times W$. However, the output of the inertial pose estimation $\bar{u}_t$ is a vector (r, x, y) of size 3. To be consistent, we encode $\bar{u}_t$ as a one-hot tensor $E(u_t)$ with dimensions $R \times W \times H$. Combining the visual and inertial estimation, the probabilistic distribution of the pose estimation is chosen as the following:
\begin{equation}
    \label{eq:threshold function2}
\def\thefootnote{a}\footnotesize
    \begin{aligned}
    p_t = 
     \begin{cases}
        v_t & \text{if $\bar{p}_t = \bar{v}_t$}\\
        E(u_t) & \text{otherwise}
    \end{cases}
    \end{aligned}
\end{equation}

Subsequently, projecting the egocentric observation onto the map coordinate is achieved using transposed convolution over $p_t$ with kernel $\acute{o}_t$ and stride 1:
\begin{equation}
    \label{eq:pieces}
\def\thefootnote{a}\footnotesize
    \begin{aligned}
     m \;\mbox{*}\; \acute{o}_t
    \end{aligned}
\end{equation}
The output $\tilde{o_t} \in \mathbb{R}^{R \times H \times W}$ represents the  allocentric map that contains observation information 

%This procedure can be conceptually explained as follows: The spatial transformation of a feature vector based on the pose can be achieved by projecting it onto the corresponding one-hot tensor using cross-correlation with a flipped direction. This process is analogous to the transposed convolution operation commonly used in deep neural networks. Compare to spatial transformation network\cite{spatialTransformer}, the convolution preserving the distribution of pose estimation within the resulting transformed feature map $\tilde{o_t}$. 

\textbf{ROI}. We create an $h \times h$ Region of Interest (ROI) around the estimated pose $\bar{p}_t$. The global map within the ROI undergoes updates while the areas outside the ROI remain unchanged. This strategy aims to mitigate errors, especially when visual pose estimation $v_t$ contains high levels of uncertainty.
% The global map within the ROI will be updated and the global map outside the ROI will remain to be the same. 
To separate the ROI from the rest of the map, an ROI mask, $M_t$ is created as the following:
\begin{equation}
    \label{eq:mask}
\def\thefootnote{a}\footnotesize
    \begin{aligned}
    M_t^{x,y} = 
     \begin{cases}
        1 & \text{if $x \in [\bar{p}_t^x-\frac{h}{2}, \bar{p}_t^x+\frac{h}{2}]$, $y \in [\bar{p}_t^y-\frac{h}{2}, \bar{p}_t^y+\frac{h}{2}]$}\\
        0 & \text{otherwise}
    \end{cases}
    \end{aligned}
\end{equation}

\textbf{Update}. Given the projected observation $\tilde{o_t}$ and the ROI mask, we update the global map $m_t$ using a convolutional LSTM model \cite{convlstm}. The model learns how to "remember" or "forget" the information in the current map, and whether the information in the incoming observation can be trusted and stored in the map. The update of the global map can be expressed in the following way 
\begin{equation}
    \label{eq:h}
\def\thefootnote{a}\footnotesize
    \begin{aligned}
    \mathnormal {m}_t = \sigma((1-M_t)\odot m_{t-1} + M_t \odot ConvLSTM(m_{t-1},\tilde{o_t}))
    \end{aligned}
\end{equation}
where $\odot$ denotes the Hadamard Product.  The entry of the map is a vector of size $L$, which represents probabilities of $L$ class labels of the foreground object located at this position. We use the ROI mask $M_t$ to select the region that needs to be updated, and retain the original value of entries outside the ROI.

\subsection{Loss function}
In the proposed SemanticSLAM, the accuracy of the constructed semantic map plays a critical role in pose estimation. An accurate global map ensures correct pose estimation. Therefore, the ConvLSTM model is trained to improve the quality of the constructed map. We design a loss function that quantifies the accumulated disparity between the constructed map $m_t$ and the ground truth map $\overline{m}_t$ over $T$ time steps as the following:

\begin{equation}
    \label{eq:loss}
\def\thefootnote{a}\footnotesize
    \begin{aligned}
    \mathcal{L} = \frac{1}{T} \sum_{t}^{T}\sum_{x,y}^{H, W} \mathcal{D}_{KL}(\overline{m}_{t,x,y}||m_{t,x,y})
    \end{aligned}
\end{equation}
where $\mathcal{D}_{KL}(\overline{m}_{t,x,y}||m_{t,x,y})$ represents the Kullback–Leibler divergence(KL) between the ground truth label and the estimated label of the map location $(x,y)$ at time $t$.

% \begin{equation}
%     \label{eq:kl}
% \def\thefootnote{a}\footnotesize
%     \begin{aligned}
%     \mathcal{D}_{KL}(Q||P)=\sum_{x\in \mathcal{X}}Q(x)log(\frac{Q(x)}{P(x)})
%     \end{aligned}
% \end{equation}

\begin{figure}[t]
\centering\includegraphics [width=0.3\textwidth]{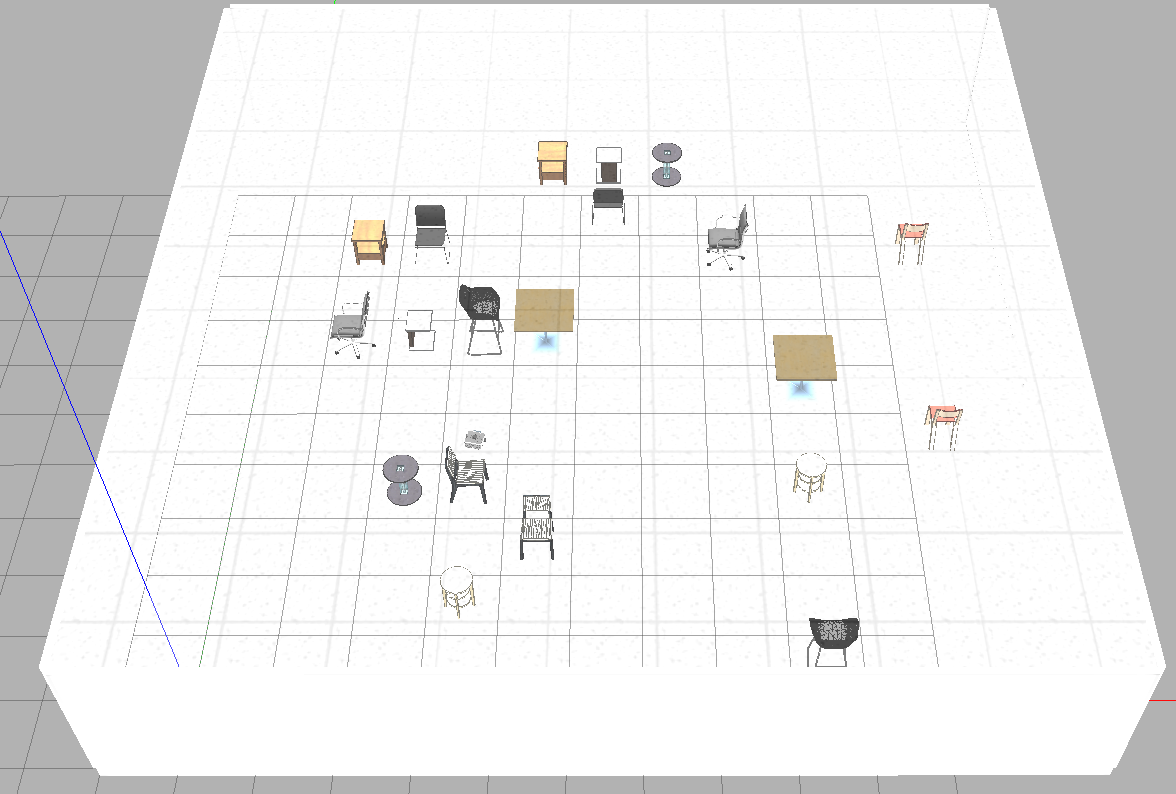}
\caption{\centering \textbf{IndoorScenes Dataset}}
\label{fig:env}
\vspace{-0.5cm}
\end{figure}

\section{Experiments}
\subsection{Experiment Setup}
Our evaluation dataset, "IndoorScenes", is a simulated indoor localization dataset. An example of the indoor scene is shown in Figure\ref{fig:env}. The environment is the indoor environment, featuring various objects randomly placed within it, generated by Gazebo\cite{gazebo}, an open-source multi-robot simulator. To simulate low-frequency sensor input, a ground robot simulator, TurtleBot3\cite{turtlebot}, captures RGB and Depth images with 640x480 resolution and a horizontal filed of view(FOV) of 90 degrees as it traverses the environment. The images are sampled at a rate slower than 1Hz and the IMU sensor contained Gaussian noise and biased noise. We ran the simulator to generate 30 different scenes. Each scene contains 3 different trajectories.

The map size is $33 \times 33$, which means the grid size is 300 millimeters. The dimension of the observation map is $11 \times 11$ and the orientation of the camera is discretized into $R=360$ levels. 

We conducted our experiment under two different settings, "Intra-Scene" and "Cross-Scene". Under the "Intra-Scene" setting, we train the model with two of the three trajectories from all 30 scenes, while test the model using the remaining trajectory from each scene. Under the "Cross-Scene" setting, the dataset was divided into two distinct sets with different scenes to ensure that the testing trajectory and training trajectory are collected from different scenes. The aim of "Cross-Scene" testing is to analyze whether our model can generalize to a new training environments. 

% We trained our model using trajectory sequences with $T=80$ inputs, and batch size of 28. 

\begin{table}[tb]
\centering
\caption{\centering \textbf{Performance comparison.} (Average Position Error(APE), Direction Error(DE))}
\label{tb:table1}
    \def\thefootnote{a}\footnotesize
    \centering
   \begin{tabular}{ |*{6}{c|}}
        \hline
         % & \multirow{2}{*}{SM} & \multirow{2}{*}{drift} & \multirow{2}{*}{allo} & Intra-Scene & Cross-Scene \\
         % & map & corr. & & &\\
         & SM & DC & Allo & Intra-Scene & Cross-Scene\\
        \hline
        DeepVO\cite{deepvo} & &  &  & 2.29, 87.31 & 3.97, 103.7  \\
        \hline
        MapNet\cite{mapnet} &  &  & \checkmark & 2.40, 106.1 & 3.90, 92.42   \\
        \hline
        ORBSLAM2\cite{orbslam2} &  & \checkmark & \checkmark & N/A & N/A   \\
        \hline
        our(visual) & \checkmark & \checkmark & \checkmark & 2.14, 49.37 & 3.24, 67.73 \\
        \hline
        our(visual-inertial) & \checkmark & \checkmark & \checkmark & \textbf{1.61, 34.64}  & \textbf{2.39, 57.11} \\
        \hline
        % ours(perfect) & Yes & Yes & Yes & \textbf{1.154m, 23.71°, 0.54\%}\\
        % \hline
    \end{tabular}
    
    \raggedleft\text{SM: semantic map, DC: drift correction, Allo: allocentric pose estimation}
\end{table}

% \begin{table}[t]
% \centering
% \caption{\centering \textbf{Performance for sequence of 80 frames, on the Cross-Scene dataset} }
% \label{tb:table3}
%     \def\thefootnote{a}\footnotesize
%     \centering
%     \begin{tabular}{|c|c|c|c|}
%     \hline
%     & APE & DE\\
%     \hline
        
%     DeepVO & 4.487 & 92.64 \\
%     \hline
%     Mapnet & 4.337 & 93.85 \\
%     \hline
%     SemanticSLAM(Monocular) & \textbf{3.98} & \textbf{75.13} \\
%     \hline
%     \end{tabular}
% \end{table}

\begin{table}[t]
\centering
\caption{\centering \textbf{Path From base model to SemanticSLAM} }
\label{tb:table2}
    \def\thefootnote{a}\footnotesize
    \centering
    \begin{tabular}{|c|c|c|c|c|}
    \hline
    & Mapnet & +Semantic & +ConvLSTM & SemanticSLAM \\
    \hline
    result&3.90, 92.42& 3.80, 80.50& 3.24, 67.73& 2.39, 57.11\\
    \hline
    \end{tabular}
    \vspace{-0.3cm}
\end{table}

Two evaluation metrics are used to measure the quality of the model. The average position error (APE) is determined by calculating the average of the Euclidean distances between the predicted and actual positions over time. The average direction error (ADE) is computed by calculating the mean orientation difference. 
% The accuracy metric represents the percentage of 'correct' pose where both the position error and direction error fall below $1m$ and $30^\circ$, respectively.

%\subsection{Comparison Baselines}
For comparison, we implemented and trained several existing models that utilize different localization techniques. 

\begin{itemize}
\item DeepVO: This is a typical deep learning-based RPR model that estimates relative camera pose through regression. The model takes in a pair of RGB images as input and produces predictions for the relative changes in translational motion $(\Delta x, \Delta y, \Delta\theta)$.
\item ORB-SLAM2: This is a keypoint-based VSLAM model. It leverages manually crafted feature descriptors to represent the images, which are then stored in a map. %The model accepts RGBD images as input and provides the pose estimation in the form of a 4x4 transformation matrix. This matrix can subsequently be converted to the pose $(x, y, \theta)$ in the world coordinate system. 
\item MapNet: This model is very similar to ours except that it extracts image features from RGB images and projects them to the observation map. Its global map also stores image features instead of semantic features. %maximum feature to the ground observation using depth images. The map update process is implemented through a Long Short-Term Memory (LSTM) model. As a result, the model generates a probability field indicating the likelihood of the agent being present at a particular position and orientation in the world coordinate system.
\end{itemize}

\subsection{Results and Analysis}
\subsubsection{Performance Comparison}
Table \ref{tb:table1} compares the average position error and direction error of these SLAM models under both Intra-Scene and Cross-Scene settings. Each test sequence is a trajectory with 30 steps. At each step the robot observes the environment and performs pose estimation. For each test, the robot begins with no prior information of the environment. We can see that the SemanticSLAM with both visual and IMU input has the best performance and the second best model is the SemanticSLAM with only visual sensors.  Although DeepVO also performs well in Intra-Scene setting, the model is overly customized to the scenes in the training set, therefore does not have good performance in generalization when applied to the Cross-Scene setting. In contrast, our algorithm learns how to construct the map  instead of the map itself, hence it can adapt to environment with different scenes. Finally, we also applied ORB-SLAM2 to our dataset, however, the algorithm could not generate any pose estimation because it could not find sufficient number of matched key points between frames.  %proficiency in map construction and pose prediction via point-based techniques, it falters in accurately predicting poses within our dataset. This inefficiency is largely due to ORB-SLAM2's pronounced dependence on continuous data for the extraction of a sufficient number of feature points for matching purposes.
% ORB-SLAM2, although proficient in map construction and pose prediction using point-based techniques, fails to accurately predict poses in our dataset. This can be attributed to ORB-SLAM2's heavy reliance on highly continual data for extracting an adequate number of feature points for matching. 
%In contrast, our model prioritizes semantic feature matching, resulting in greater reliability and robustness when dealing with low-frequency sensor data. Furthermore, our model outperforms MapNet in both position and orientation estimation, as evidenced by the results. 

\begin{figure}[t]
\centering\includegraphics [width=0.25\textwidth]{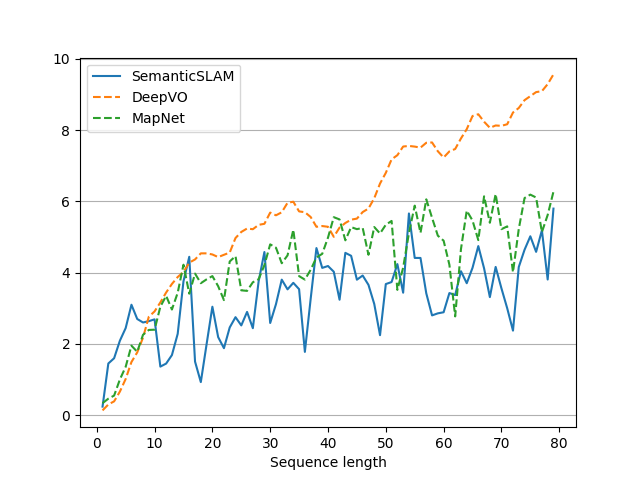}
%       \begin{subfigure}[b]{0.23\textwidth}
%         \includegraphics[width=\textwidth]{./images/apeloss_new.png}
%       \end{subfigure}
%       \hfill
%       \begin{subfigure}[b]{0.23\textwidth}
%         \includegraphics[width=\textwidth]{./images/deloss_new.png}
%       \end{subfigure}
      \caption{\centering \textbf{Localization performance over time.} Our model (green solid) vs. reference models (dashed lines)}  
    \label{fig:apede} 
    \vspace{-0.3cm}
\end{figure}

%We also measure the performance in the long-term sequence of 80 frames in Table \ref{tb:table3}, which is much longer than the training sequence of 30 frames. We verified SemanticSLAM monocular model achieves better localization result for long sequences. The semantic map seems to be more robust than feature based method in the long term.
Figure \ref{fig:apede} illustrates the localization error over time for various methods. It shows that with SemanticSLAM(visual) the error accumulation is significantly slowed down.  

% \begin{table}[htb]
% \centering
% \caption{\centering \textbf{Path From similar model to SemanticSLAM.} }
% \label{tb:table2}
%     \def\thefootnote{a}\footnotesize
%     \centering
%     \begin{tabular}{c|c|}
%      Model & Result \\
%     \hline
%     Mapnet & 3.904, 92.42°\\
%     +semantic & 3.807, 80.50°\\
%     +ConvLSTM & 3.242, 67.73°\\
%     +IMU crosscheck & 2.399, 57.11
%     \end{tabular}
% \end{table}

% \begin{table*}[htb]
% \centering
% \caption{\centering \textbf{Path From similar model to SemanticSLAM.} }
% \label{tb:table2}
%     \def\thefootnote{a}\footnotesize
%     \centering
%     \begin{tabular}{c|c|cc|c}
%    % \begin{tabular}{{c|}|{c|}|{c}{c}|{c|}}
            
%           & MapNet & SemanticSLAM(v1) & SemanticSLAM(v2) & SemanticSLAM(final) \\
%         \hline
%         Semantic?  & & $\checkmark$ & $\checkmark$ & $\checkmark$\\
%         ConvLSTM? & & &$\checkmark$ &$\checkmark$\\ 
%         IMU crosscheck?  & & & &$\checkmark$\\
%         \hline
%         Result & 3.904m, 92.42°, 0.56\% & 3.807m, 80.50°, 4.34\% & 3.242m, 67.73°, 3.69\% & \textbf{2.399, 57.11 , 35.00\%}\\
%     \end{tabular}
% \end{table*}

\subsubsection{Ablation Study}

The SemanticSLAM is inspired by MapNet. Compared to MapNet, the following enhancements were adopted, (1) semantic features instead of image features were extracted from the observation and stored in the global map, a new feature projection algorithm was developed; (2) the map update is achieved using a ConvLSTM instead of a normal LSTM; (3) we bootstrap the pose estimation by cross-checking the visual and inertial information. In Table \ref{tb:table2}, we show how much improvement each of these enhancement techniques could bring. As we can see, by using semantic features and adopting new feature projection algorithms, we can reduce the APE by 2\% and DE by 14\%. By adopting the ConvLSTM model, we can further reduce the APE and DE by 17\% and 19\% respectively. Finally by utilizing the inertial information, the APE and DE can be further reduced by 35\% and 18\% respectively.

%starting with MapNet as the base model, we gradually introduce each technique to showcase the resulting improvements. The first notable technique is the inclusion of semantic features. This aids in reducing the error rate by focusing on the region that is most likely to contain significant information for localization, thus delivering a more reliable feature compared to traditional pure CNN backbone. Besides, the detailed categorization of each pixel contributes to distinguish between different types of objects and spatial features within a scene. This makes our approach robust against cluttered and diverse environments, ensuring the creation of detailed and accurate maps. The projection procedure enables a refined understanding of space occupancy and provides a more informative measure of the object’s spatial distribution and scale. 

% This addition is crucial as it leverages the distinctive and viewpoint-invariant nature of semantic classes, thereby facilitating accurate camera localization. 
\subsubsection{Map Construction}
Next we will demonstrate the error correction capability of SemanticSLAM during map construction. The baseline map construction algorithm is a heuristic that update each grid of the global map in a leaky-integrate manner: 
\begin{equation}
    \label{eq:heuristic}
\def\thefootnote{a}\footnotesize
    \begin{aligned}
    \mathnormal{m}_t(x,y) = (1-\alpha) \times \mathnormal{m}_{t-1}(x,y) + \alpha \times \tilde{o_t}(x,y)  \quad \text{if $\tilde{o_t}(x,y) > 0$}\\
    \end{aligned}
\end{equation}
%Our second technique utilizes ConvLSTM to handle uncertainties in the map, especially those caused by visual noise from the semantic processing stage (referenced in Section \ref{sec:v_noise}) and drift errors from the IMU. 
where $\alpha$ varies from 0.1, 0.3 to 0.7 and 1. It controls the leakage of the old information in the map and map update speed.  Table \ref{tb:table4}  compares the MSE error of the maps constructed using our method and using the heuristic method. To have a fair comparison, we assume that we have perfect knowledge of the camera pose, and the only error is from the observation. The errors in the observation may come from 3 sources. The first type of error is the feature extraction and projection error. The object detection, classification and segmentation may not be perfect and we may project the feature to a wrong position in the observation map $o_t$. The second type of error is the obstruction error. Even if the feature extraction and projection are perfect, we still may miss some objects in the scene due to obstruction by other objects. The third type of error is rotation error. Even if there is no objection, after rotating the $o_t$ to generate the allocentric observation candidate $\acute{o}_t$, we may introduce interpolation error. To feature these different errors, we created three different type of observation inputs. The "Real Camera" input contains all 3 types of errors, the "Obstructed" input contains the rotation error and obstruction error, and the "Ideal" input contains only the rotation error. As shown in Table \ref{tb:table4}, the map constructed by our algorithm has much lower MSE error compared to the heuristic algorithms.

\begin{figure}[t]
\centering\includegraphics [width=0.25\textwidth]{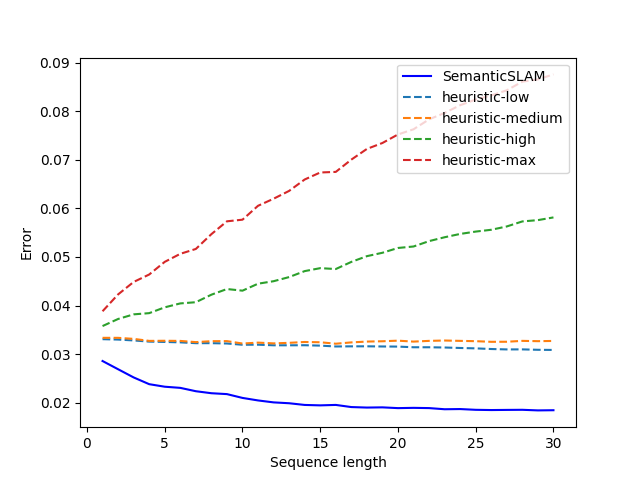}
      \caption{\centering\textbf{Map construction loss over time}}
    \label{fig:heuristic} 
    \vspace{-0.5cm}
\end{figure}

Figure \ref{fig:heuristic} shows how the map construction error changes as the mission goes on. Using our map construction, the map error reduces with the increase of time steps. Such trend is not observed with the heuristic algorithms.

%These heuristic methods transform semantic observation into the map coordinate using the ground-truth pose. Then the semantic observation are integrated into the map, scaled by a update rate, which controls the balance between retaining the previous value and incorporating the new input during each update. The table assesses three different qualities of semantic observation, from poor to good. The results show that ConvLSTM, by capturing spatial-temporal features from local neighbors and past states, can correct errors during map update, ensuring the map accuracy for camera localization. Example results are shown in Figure\ref{fig:heuristic}.

%In addition to the aforementioned techniques, the IMU cross-checking plays a significant role in preventing accumulated errors. It achieves this by cross-checking visual input with the IMU, allowing the inertial pose to guide the map update process when visual image quality is low. Moreover, the ROI mask effectively decouples the computational cost from the scene scale. It accomplishes this by preserving temporal-spatial information within the designated region while preventing information loss outside of the region within the LSTM. Consequently, the model exclusively focuses on updating the Region of Interest (ROI) within the map, deviating from the conventional approach of updating the entire allocentric map. 
% as observed in prior works \cite{neuralslam,neuralmap,mapnet}.

\begin{table}[t]
\centering
\caption{\centering \textbf{Map construction loss of our model compared to heuristic methods.} MSE error and standard deviation(x100)}
\label{tb:table4}
    \def\thefootnote{a}\footnotesize
    \centering
   \begin{tabular}{ |*{4}{c|}}
        \hline
        Condition & Real camera & Obstructed & Ideal\\
        \hline
        our & \textbf{2.51$\pm$0.21} &\textbf{ 2.42$\pm$0.21} & \textbf{2.05$\pm$0.29}\\
        \hline
        heuristic(0.1) & 2.90$\pm$0.06 & 3.21$\pm$0.04& 3.17$\pm$0.05\\
        \hline
        heuristic(0.3) & 2.57$\pm$0.12& 3.18$\pm$0.10& 3.26$\pm$0.15\\
        \hline
        heuristic(0.7) & 2.57$\pm$0.28& 3.65$\pm$0.26& 4.77$\pm$0.41\\
        \hline
        heuristic(1.0) &2.94$\pm$0.45 & 4.35$\pm$0.41& 6.65$\pm$0.59\\
        \hline
    \end{tabular}
    \vspace{-0.3cm}
\end{table}

\begin{figure}[t]
\includegraphics [width=0.48\textwidth]{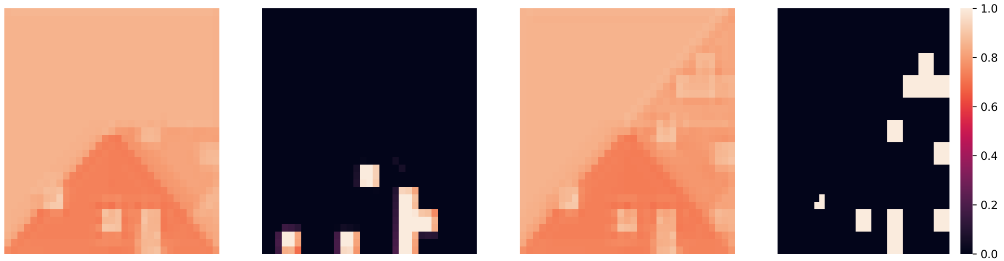}
\caption{\textbf{Visualization of the Map Update Process}. From left to right: Original global map $m_{t-1}$, Semantic observation $o_{t}$, Updated global map $m_{t}$, and Ground-truth.}  
\label{fig:obs_map} 
\vspace{-0.5cm}
\end{figure}

% \begin{figure}[t]
% \centering\includegraphics [width=0.3\textwidth]{./images/trajectory.png}
% \caption{Cross-Scene Estimated trajectory (blue) and ground-truth (red)}  
% \label{fig:trajectory} 
% \end{figure}

%\subsubsection{Observation and Map Prediction Visualization}
%In our last section, we provide a qualitative demonstration of our model by visualizing the semantic map predictions, as shown in Figure \ref{fig:obs_map}. The figure presents feature embeddings where each cell in the state initially presents a probability distribution vector of size $L$. This vector reflects the confidence level of the cell being populated by an object. To enhance the visualization of feature embeddings, we combine all $L$ semantic layers of each cell, excluding the non-object layer. Consequently, the value of each cell represents the  the probability of an object at the corresponding location. Notably the unexplored top-left area of the map appears with identical colors, indicating an equal probability for all categories, including non-objects. 

 Figure\ref{fig:obs_map} illustrate the map construction process. We pick one of the $L$ channels of the global map $m_t$ and egocentric observation map $o_t$ and show them in the figure. The updated map and the ground truth map is also given. The brightness of the pixel indicate the confidence that this grid cell is occupied.
 %, our model adeptly integrates the observations into the map, generating predictions that are in strong concordance with the ground truth labels. Moreover, uncertainties present in the initial map are rectified as the model absorbs and processes more information.

\section{Conclusions}
This work presents a pioneering semantic learning-based SLAM system that significantly enhances localization accuracy, particularly for a system with infrequent observation environment. The constructed map offers a comprehensive semantic-level representation of the environment that could be utilized for navigation planning or shared among robots. It also proved interpretable information for human users. Our research demonstrates the practicality and potential of incorporating semantic features into SLAM. %On the other hand, our method is restricted to the scale of the 
% Moving forward, our future directions include developing a more resilient and adaptable mechanism capable of effectively balancing IMU and visual inputs, thereby enabling online training across a wide range of similar scenarios. Additionally, our aim is to train a model that can accurately predict areas that are either occluded or unexplored, further enhancing the mapping capabilities of the system.

\section{Acknowledgements}
This research was partially supported by NSF IUCRC ASIC center (CNS-1822165) and NSF award CNS-2148253.

\bibliographystyle{plain}
\bibliography{reference}

\begin{thebibliography}{10}

\bibitem{selfvio}
Yasin Almalioglu et~al.
\newblock Selfvio: Self-supervised deep monocular visual-inertial odometry and depth estimation.
\newblock {\em CoRR}, abs/1911.09968, 2019.

\bibitem{dynaslam}
Berta Besc{\'{o}}s et~al.
\newblock Dynslam: Tracking, mapping and inpainting in dynamic scenes.
\newblock {\em CoRR}, abs/1806.05620, 2018.

\bibitem{liftslam}
Hudson Martins~Silva Bruno and Esther~Luna Colombini.
\newblock {LIFT-SLAM:} a deep-learning feature-based monocular visual {SLAM} method.
\newblock {\em CoRR}, abs/2104.00099, 2021.

\bibitem{orbslam3}
Carlos Campos et~al.
\newblock {ORB-SLAM3:} an accurate open-source library for visual, visual-inertial and multi-map {SLAM}.
\newblock {\em CoRR}, abs/2007.11898, 2020.

\bibitem{neuralslam}
Devendra~Singh Chaplot, Dhiraj Gandhi, Saurabh Gupta, Abhinav Gupta, and Ruslan Salakhutdinov.
\newblock Learning to explore using active neural {SLAM}.
\newblock {\em CoRR}, abs/2004.05155, 2020.

\bibitem{vinet}
Ronald Clark, Sen Wang, Hongkai Wen, Andrew Markham, and Niki Trigoni.
\newblock Vinet: Visual-inertial odometry as a sequence-to-sequence learning problem.
\newblock {\em CoRR}, abs/1701.08376, 2017.

\bibitem{eichenbaum1999hippocampus}
Howard Eichenbaum, Paul Dudchenko, Emma Wood, Matthew Shapiro, and Heikki Tanila.
\newblock The hippocampus, memory, and place cells: is it spatial memory or a memory space?
\newblock {\em Neuron}, 23(2):209--226, 1999.

\bibitem{mapnet}
Joao~F. Henriques and Andrea Vedaldi.
\newblock Mapnet: An allocentric spatial memory for mapping environments.
\newblock In {\em 2018 IEEE/CVF Conference on Computer Vision and Pattern Recognition}, pages 8476--8484, 2018.

\bibitem{herweg2018spatial}
Nora~A Herweg and Michael~J Kahana.
\newblock Spatial representations in the human brain.
\newblock {\em Frontiers in human neuroscience}, 12:297, 2018.

\bibitem{spatialTransformer}
Max Jaderberg, Karen Simonyan, Andrew Zisserman, and Koray Kavukcuoglu.
\newblock Spatial transformer networks.
\newblock {\em CoRR}, abs/1506.02025, 2015.

\bibitem{posenet}
Alex Kendall, Matthew Grimes, and Roberto Cipolla.
\newblock Posenet: A convolutional network for real-time 6-dof camera relocalization.
\newblock {\em IEEE international conference on computer vision}, pages 2938--2946, 2015.

\bibitem{SegAnything}
Alexander Kirillov, Eric Mintun, Nikhila Ravi, Hanzi Mao, Chloe Rolland, Laura Gustafson, Tete Xiao, Spencer Whitehead, Alexander~C. Berg, Wan-Yen Lo, Piotr Doll{\'a}r, and Ross Girshick.
\newblock Segment anything.
\newblock {\em arXiv:2304.02643}, 2023.

\bibitem{gazebo}
Nathan Koenig and Andrew Howard.
\newblock Design and use paradigms for gazebo, an open-source multi-robot simulator.
\newblock In {\em IEEE/RSJ International Conference on Intelligent Robots and Systems}, pages 2149--2154, Sendai, Japan, Sep 2004.

\bibitem{sift}
David~G. Lowe.
\newblock Distinctive image features from scale-invariant keypoints.
\newblock {\em Int. J. Comput. Vision}, 60(2):91--110, November 2004.

\bibitem{deepvo}
Vikram Mohanty, Shubh Agrawal, Shaswat Datta, Arna Ghosh, Vishnu~Dutt Sharma, and Debashish Chakravarty.
\newblock Deepvo: A deep learning approach for monocular visual odometry.
\newblock {\em arXiv preprint arXiv:1611.06069}, 2016.

\bibitem{orbslam2}
Raul Mur{-}Artal and Juan~D. Tard{\'{o}}s.
\newblock {ORB-SLAM2:} an open-source {SLAM} system for monocular, stereo and {RGB-D} cameras.
\newblock {\em CoRR}, abs/1610.06475, 2016.

\bibitem{yolov3}
Joseph Redmon and Ali Farhadi.
\newblock Yolov3: An incremental improvement.
\newblock {\em CoRR}, abs/1804.02767, 2018.

\bibitem{maskfusion}
Martin R{\"{u}}nz and Lourdes Agapito.
\newblock Maskfusion: Real-time recognition, tracking and reconstruction of multiple moving objects.
\newblock {\em CoRR}, abs/1804.09194, 2018.

\bibitem{superglue}
Paul{-}Edouard Sarlin et~al.
\newblock Superglue: Learning feature matching with graph neural networks.
\newblock {\em CoRR}, abs/1911.11763, 2019.

\bibitem{lessismore}
Greg Schohn and David~A. Cohn.
\newblock Less is more: Active learning with support vector machines.
\newblock In {\em ICML}, 2000.

\bibitem{convlstm}
Xingjian Shi et~al.
\newblock Convolutional {LSTM} network: {A} machine learning approach for precipitation nowcasting.
\newblock {\em CoRR}, abs/1506.04214, 2015.

\bibitem{turtlebot}
{Tully Foote, Melonee Wise}.
\newblock Turtlebot, 2011.

\end{thebibliography}
\end{document}